\documentclass[10pt]{article} 
\usepackage[preprint]{rlc}

\usepackage{amssymb}            
\usepackage{mathtools}          
\usepackage{mathrsfs}           
\mathtoolsset{showonlyrefs}     
\usepackage{graphicx}           
\usepackage{subcaption}         
\usepackage[space]{grffile}     
\usepackage{url}             
\usepackage{xurl}
\usepackage{rotating}
\usepackage{enumitem}
\usepackage{xcolor}
\usepackage{etoolbox}
\usepackage{placeins}
\makeatletter
\def\fps@figure{!htbp}
\def\fps@table{!htbp}
\makeatother
\usepackage{tabularx, booktabs, multirow}
\newcolumntype{L}{>{\raggedright\arraybackslash}X}

\definecolor{rqanswerbg}{HTML}{F3F6FA}
\definecolor{rqanswerborder}{HTML}{CFD8E3}
\definecolor{auditnotebg}{HTML}{F7F4EE}
\definecolor{auditnoteborder}{HTML}{D9CDBB}
\newcommand{\rqanswer}[1]{%
    \begin{center}
    \begingroup
    \setlength{\fboxsep}{6pt}%
    \fcolorbox{rqanswerborder}{rqanswerbg}{%
        \begin{minipage}{0.91\linewidth}
        \small\textbf{RQ answer.} #1
        \end{minipage}}%
    \endgroup
    \end{center}
}

\newtoggle{final}
\togglefalse{final} 

\def\email{\small\ttfamily}
\title{\centering 
SkillJuror: Measuring How Agent Skill Organization Changes Runtime Behavior
}

\author{
\parbox{\textwidth}{
\centering
Zhiyu Chen$^{1,2,\dagger}$, 
Zihan Guo$^{2,3,\dagger}$, 
Bo Huang$^{2,4}$, 
Bingwei Lu$^{4}$, 
\\
Jianghao Lin$^{4*}$, 
Yuanjian Zhou$^{2*}$,
Weinan Zhang$^{2,4*}$
}
\vspace{0.5em} 
\\
$^{1}$ Tongji University \quad$^{2}$ Shanghai Innovation Institute
\\
$^{3}$ Sun Yat-sen University \quad$^{4}$ Shanghai Jiao Tong University 
\\
$^{\dagger}$ Equal contribution.\quad$^{*}$~Corresponding author.
\\
\email{2354271@tongji.edu.cn, guozh29@mail2.sysu.edu.cn, wnzhang@sjtu.edu.cn}
}

\begin{document}

\maketitle

\begin{abstract}

Agent Skills augment large language model (LLM) agents with procedural knowledge at inference time, but current benchmarks rarely distinguish what a Skill says from how it is organized. We study this distinction through \emph{Progressive Disclosure}, where a concise root file points agents to supporting resources on demand, and compare it with a normalized flat baseline. We present \textbf{SkillJuror}, a framework for evaluating Skill writing paradigms through semantically controlled variants, matched multi-trial evaluations, and trajectory evidence while holding task knowledge fixed. In an 82-task SkillsBench study, \emph{Progressive Disclosure} changes runtime behavior before aggregate outcomes: distinct Skill resources touched per trajectory rise from 1.18 to 3.85, and effective uptake events rise from 1.33 to 3.92. It also yields 17 additional verifier-passing trials out of 410 matched trials (+4.1\%) over the normalized flat baseline. The benefit is task-dependent. \emph{Progressive Disclosure} helps when supporting resources guide implementation, checking, or repair, but is weaker when success hinges on exact output conventions, numerical thresholds, or long artifact-generation pipelines. These results show that Skill organization is not mere presentation: it can change how agents search and apply procedural knowledge, while outcome gains depend on whether the exposed resources are actionable for the task. Code is available at \url{https://github.com/zhiyuchen-ai/skill-juror}.

\vspace{10pt}
\end{abstract}

\section{Introduction}

Agent Skills are becoming a practical mechanism for giving LLM agents task-specific procedural knowledge at inference time.
A Skill is a self-contained navigable runtime artifact rather than a static prompt.
An agent reads a root \texttt{SKILL.md}, opens supporting references, inspects templates, and may invoke helper scripts while working on a task, as described by the specification\footnote{\url{https://agentskills.io/specification}.}.
Since agents autonomously choose which resources to open and which helpers to invoke, Skill organization governs runtime trajectories instead of simply formatting documentation.

Once Agent Skills are adopted, an immediate practical question arises: \emph{how should they be written?} 
Published style guides advocate paradigms like progressive disclosure, modular organization, and scriptization~\citep{anthropic2025agentskills}. 
While these recommendations are intuitive, rigorous empirical validation remains elusive. 
Existing evidence mostly focuses on adoption, comparing whether Skills are present, absent, or self-generated, rather than how the same knowledge is organized within a Skill~\citep{li2026skillsbench}. 
Furthermore, natural Skill collections inevitably confound core knowledge coverage, individual author styles, and structural organization. 
Consequently, a better-performing Skill may owe its advantage to superior organization, to richer task knowledge, or to both. 

This attribution gap makes authoring recommendations difficult to evaluate without controlled, task-knowledge-matched variants. 
Addressing this challenge is critical because a knowledge-agnostic organization paradigm, if effective across diverse domains, would suggest that Skill layout can systematically reshape agent runtime behavior independently of task-specific content coverage. This would make organization itself a controllable mechanism for guiding how agents retrieve, apply, and reuse procedural knowledge during execution.
Isolating organization from knowledge allows us to treat structural layout as an experimentally isolated runtime intervention, establishing a scalable foundation for controlled Skill evaluation.

This paper isolates Skill organization as a controlled intervention at runtime.
Our focus is \emph{Progressive Disclosure}, the practice of keeping the root file concise and pointing agents to supporting resources on demand.
Holding procedural knowledge constant, this contrast asks whether reorganizing a flat, self-contained instruction into an on-demand resource bundle changes how the agent behaves.
\emph{Progressive Disclosure} is therefore a behavioral intervention whose outcome effects may vary by task.
A shorter root may reduce cognitive load at entry and encourage targeted reference access, while its benefits depend on whether agents access and apply the supporting resources that contain task-critical knowledge.

We introduce \textbf{SkillJuror}, an evaluation framework that isolates the behavioral and outcome effects of Skill organization by factoring out interfering variables like knowledge coverage and author style.
To disentangle organization from task knowledge, SkillJuror constructs controlled, semantically matched Skill variants from a shared source while preserving identical knowledge such as operational obligations, constraints, and helpers. 
By executing these knowledge-matched pairs under identical, reproducible runtime environments, the framework correlates macro-level pass/fail outcomes with granular trajectory evidence, such as resource access breadth and Effective Resource Uptake (ERU). 
This dual-layered measurement allows us to systematically audit whether and how a structural layout alteration modifies an agent's runtime reasoning and execution path.

\begin{figure*}[t]
    \centering
    \includegraphics[width=\textwidth]{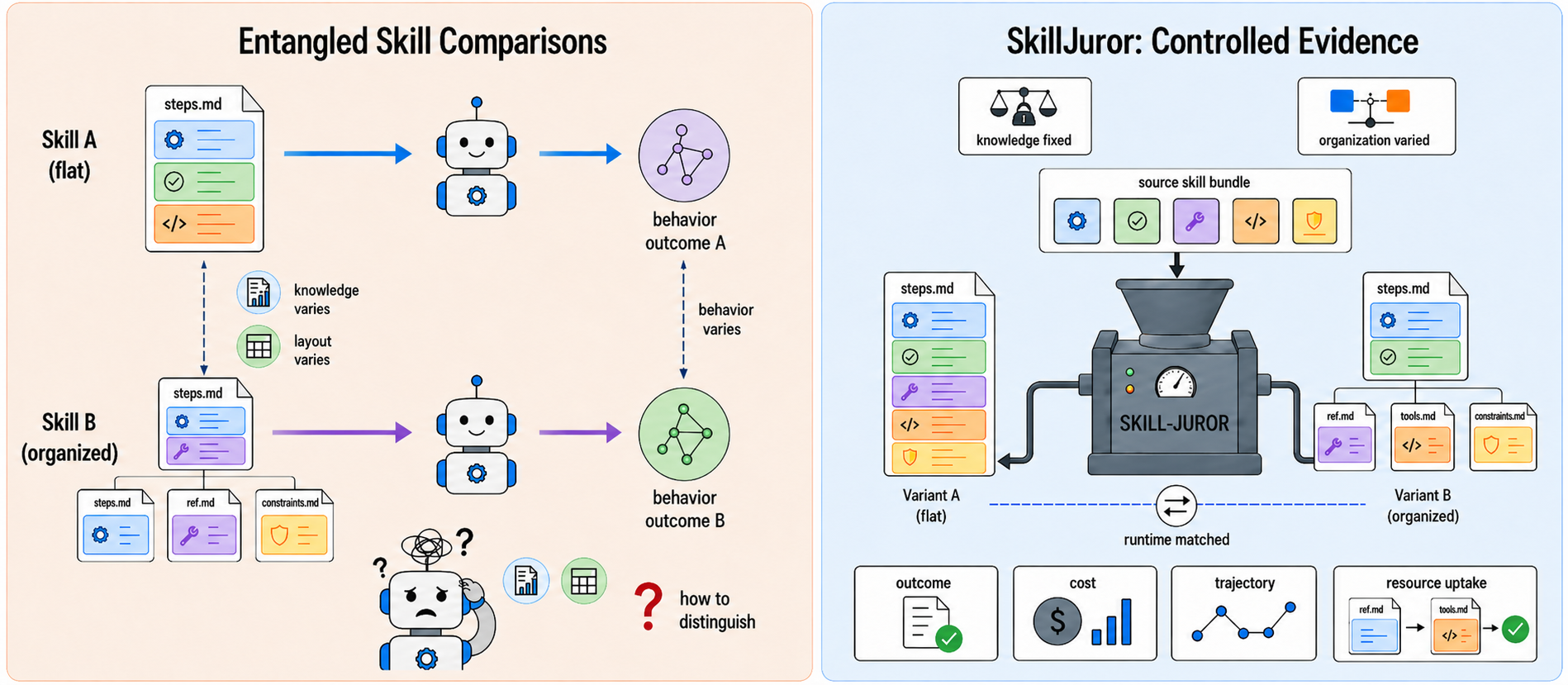}
    \caption{\textbf{From entangled Skill comparisons to controlled runtime evidence.}
    Natural Skill comparisons can vary task knowledge, artifact organization, and runtime behavior simultaneously, making outcome differences hard to attribute.
    SkillJuror instead constructs knowledge-matched variants that vary organization under matched runtime conditions, then pairs verifier outcomes with process evidence for cost, trajectory shape, and resource uptake.}
    \label{fig:intro-skill-organization}
\end{figure*}

This work makes three contributions:

\begin{enumerate}[leftmargin=2em]
    \item \textbf{The first controlled evaluation of Agent Skill organization.}
    We are, to our knowledge, the first to treat Skill organization as an experimentally isolated runtime variable, comparing task-knowledge-matched variants rather than naturally different Skills to decouple content from structure.

    \item \textbf{SkillJuror: a framework for controlled Skill-variant evaluation.}
    We introduce an audited construction and matched-runtime evaluation pipeline that preserves task knowledge while varying organization, links outcome differences to trajectory-level evidence, and enables controlled comparisons of Skill-writing paradigms.

    \item \textbf{Empirical evidence that organization changes behavior before outcomes.}
    We conduct an 82-task Progressive Disclosure study and find large shifts in resource access and effective uptake (from 1.33 to 3.92 events per trajectory). While these behavioral changes lead to a modest aggregate pass-rate gain (+4.1\%), the resulting outcome effects remain highly task-dependent.
\end{enumerate}

\section{Related Work}
\label{sec:related-work}

\subsection{Agent Skills as Runtime Procedural Artifacts}
\label{sec:rw-skills}

Agent Skills encapsulate reusable procedural knowledge as agent-facing artifacts rather than as one-off prompts. The Agent Skills specification defines a Skill as a directory centered on a \texttt{SKILL.md} file, with optional scripts, references, assets, and routing metadata. Anthropic's guidance further frames Skills around progressive disclosure: metadata supports selection, the root file provides the entry point, and supporting resources are loaded on demand~\citep{anthropic2025agentskills}. Recent procedural-memory work similarly treats skills as runtime units with activation and execution conditions, as opposed to passively retrieved text~\citep{mi2026skill}.

This view makes Skill authoring an organization problem. The same task knowledge can be exposed as a flat instruction bundle or as a concise root with supporting files. Existing specifications describe such choices as best practices, but they do not empirically test whether organization itself changes downstream behavior under controlled task semantics.

\subsection{From Skill Adoption Benchmarks to Controlled Skill-Variant Comparison}
\label{sec:rw-skillsbench}

The most closely related work evaluates whether providing Skills helps agents solve tasks. SkillsBench~\citep{li2026skillsbench} compares no-Skill, curated-Skill, and self-generated-Skill conditions over verifiable tasks, showing that curated Skills can improve pass rate but that effects are heterogeneous and self-generated Skills can be ineffective or even harmful. SWE-Skills-Bench extends this adoption question to repository-based software engineering tasks~\citep{han2026swe}, while work on larger Skill libraries shows that failures may arise from trajectory-level Skill-selection errors, not just context overhead~\citep{song2026more}.

Other benchmarks study how Skills are learned, generated, or verified. SkillLearnBench evaluates continual skill-learning methods~\citep{zhong2026skilllearnbench}; SkillGenBench evaluates pipelines that distill executable Skills from repositories or documents~\citep{zhou2026skillgenbench}; and SkillGen synthesizes auditable Skills from successful and failed trajectories~\citep{ma2026skillgen}. These efforts broaden the source of Skills, but their unit of analysis remains adoption, acquisition, or generation quality. SkillJuror instead compares variants of the same task Skill while holding task scope, helper availability, workflow obligations, and output contracts fixed, shifting the evaluation grain from Skill availability to controlled Skill-variant comparison.

\subsection{Trajectory- and Resource-Aware Agent Evaluation}
\label{sec:rw-trajectory}
\label{sec:rw-judge}

Agent evaluation has increasingly moved beyond final task success. Outcome-centric benchmarks and surveys provide scalable pass-rate or reward signals, but they often hide why an agent succeeded or failed~\citep{luo2025large}. Trajectory-aware benchmarks address this gap by evaluating intermediate behavior, including tool-use diagnostics, reasoning trajectories, and step-level process quality~\citep{he2025traject,kim2025beyond,fan2026agentprocessbench}. Cost-aware evaluation further argues that success should be interpreted together with resource consumption~\citep{erol2025cost}. To properly isolate these behavioral factors, contemporary harness engineering serves as a unifying runtime layer to eliminate uncontrolled environment variation~\citep{zhou2026externalization}, allowing SkillJuror to hold the execution substrate strictly invariant for controlled attribution.

Within this controlled framework, SkillJuror uses trajectory evidence to complement verifier outcomes by showing whether changing Skill layout changes runtime resource access and uptake. To evaluate nuanced resource-use behaviors within these trajectories, SkillJuror adopts an LLM-as-a-judge approach~\citep{gu2024survey,shi2025judging}. This choice is motivated by the need for semantic judgment when capturing effective resource uptake, allowing us to leverage LLM-assisted labels as valuable process evidence that programmatic metrics alone would miss.

\subsection{Constrained Optimization of Agent-Facing Artifacts}
\label{sec:rw-optimization}

A related line of work automatically improves agent-facing artifacts. Prompt-sensitivity work shows that even meaning-preserving formatting choices can change model behavior~\citep{sclar2024quantifying}, while prompt optimization searches over instructions, examples, or LM-program components to maximize validation performance~\citep{zhou2022large,yang2024large}. Skill-level optimization similarly edits Skill documents using rollout feedback and held-out validation~\citep{yang2026skillopt}. Tool-interface and procedural-memory systems improve how agents call APIs or accumulate reusable behavior~\citep{schick2023toolformer,song2023restgpt,shinn2023reflexion,zhao2024expel,wang2023voyager}.

These methods search for better artifacts or accumulate new behavior from experience. SkillJuror uses transformation differently by constructing controlled counterfactuals for evaluation. This transformation is designed to isolate whether the same task knowledge behaves differently when organized under a specified writing paradigm. In this paradigm-constrained setting, runtime differences can be interpreted as effects of organization rather than as unconstrained artifact optimization.

\section{Method}
\label{sec:method}

\subsection{Evaluation Problem}

SkillJuror evaluates whether changing a Skill's organization changes agent behavior while holding task knowledge fixed. Let $\mathcal{T}$ denote the task set, where each task $T$ consists of an instruction, an initial environment, and a verifier. For each task $T$ and runtime condition $c$, $\mathcal{B}_{T,c}$ denotes the Skill bundle exposed to the agent. A bundle may contain metadata, a root \texttt{SKILL.md}, executable helpers, and non-executable support material. The controlled comparison uses only bundles that pass the same construction protocol and are accepted as preserving task semantics. In the main run, the primary writing-paradigm contrast is the normalized flat baseline (Baseline) versus the Progressive Disclosure (PD) variant.

Each trial $r_{T,c,j}$ is an independent execution of task $T$ under condition $c$, and repeated trials form $\mathcal{R}_{T,c}$. For each evaluation dimension $d$, such as outcome, efficiency, paradigm realization, or routing quality, SkillJuror computes a task-level summary $A_d(T,c)$ instead of collapsing all evidence into one score. The main within-task Progressive Disclosure contrast is $\Delta_d(T)=A_d(T,\mathrm{PD})-A_d(T,\mathrm{Base})$, with benchmark- and group-level results aggregating these task-level comparisons.

\subsubsection{Overall Pipeline}

\begin{figure*}[!htbp]
    \centering
    \includegraphics[width=\textwidth]{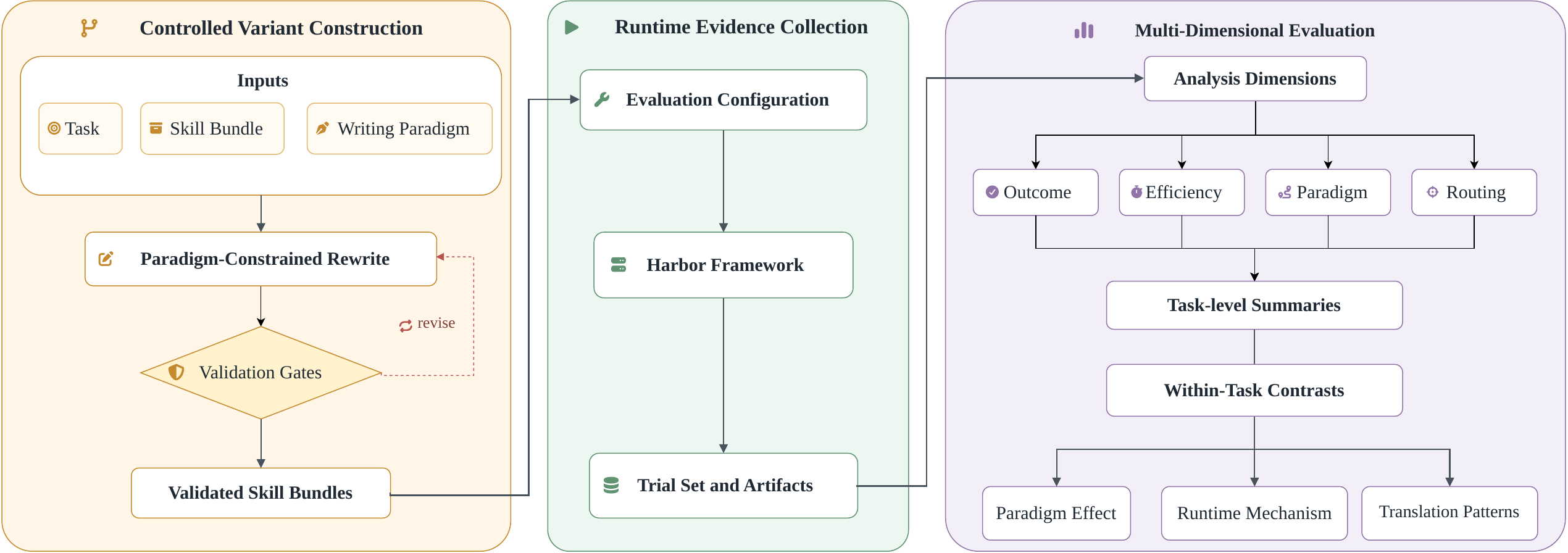}
    \caption{\textbf{Overview of the SkillJuror method pipeline.} Tasks, source Skill bundles, and target writing paradigms are transformed into accepted controlled Skill variants under knowledge-preservation checks. Accepted variants are executed under fixed runtime settings and mapped into four analysis dimensions: outcome, efficiency, paradigm realization, and routing quality. Failed construction candidates loop back for revision before entering runtime evaluation.}
    \label{fig:overview}
\end{figure*}

SkillJuror follows the three-stage pipeline in Figure~\ref{fig:overview}. Taking a source Skill bundle and a target writing paradigm as input, it rewrites the Skill under a knowledge-preservation constraint so that accepted variants differ primarily in organization rather than in knowledge content. It then executes\footnote{\href{https://github.com/harbor-framework/harbor}{Harbor} provides the containerized sandbox for both Skill artifact generation and task execution.} each task--condition pair in repeated, matched trials and maps the resulting artifacts into dimension-specific summaries for outcome, efficiency, paradigm realization, and routing quality.

\subsection{Controlled Skill Variant Comparison}

\begin{figure*}[!htbp]
    \centering
    \includegraphics[width=\textwidth]{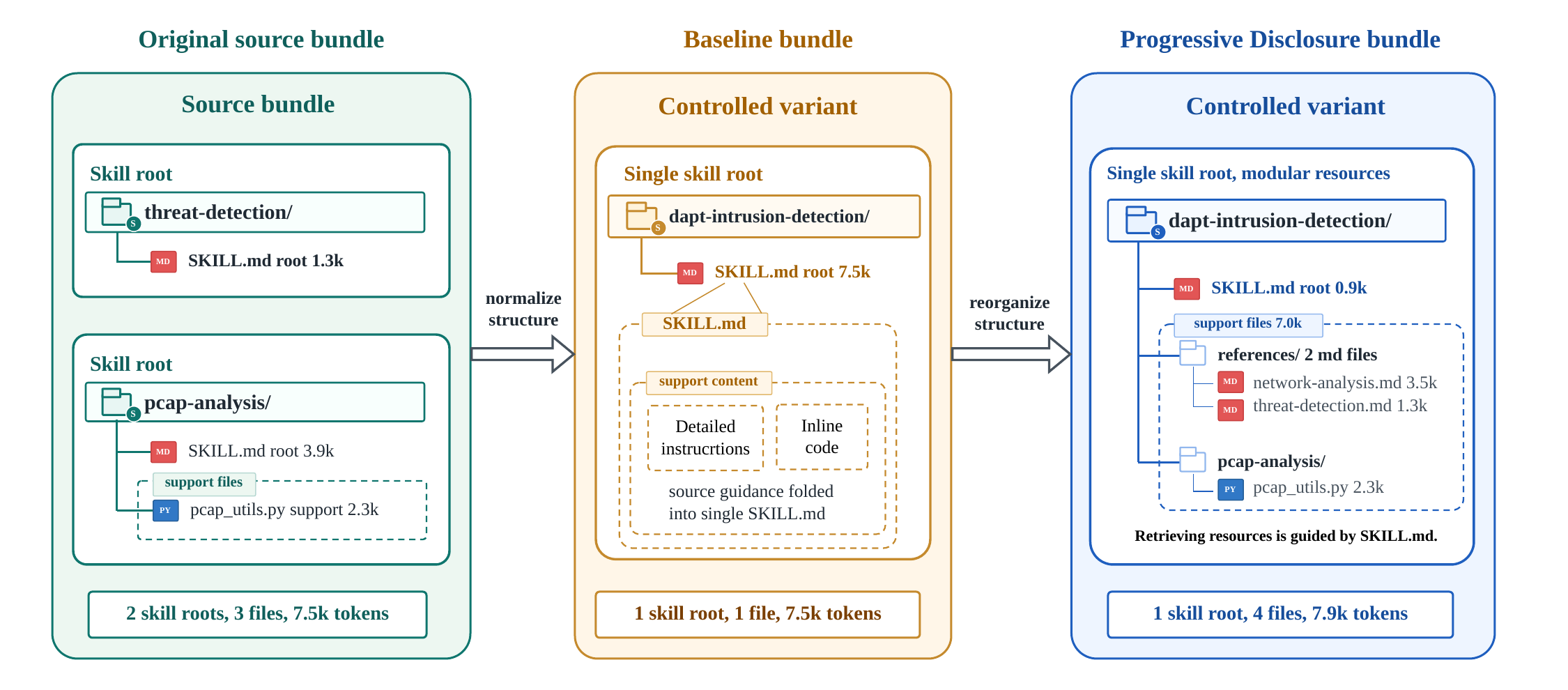}
    \caption{\textbf{Example controlled Skill-bundle transformation.} For illustration, we select a SkillsBench source bundle with multiple Skill roots and show how it is converted into the Baseline artifact, then reorganized into a Progressive Disclosure bundle with a concise entry point and on-demand support files. The example is drawn from \texttt{dapt-intrusion-detection}; token counts are approximate tokenizer counts over file contents and are shown only to indicate relative size.}
    \label{fig:controlled-skill-bundle-example}
\end{figure*}

Directly evaluating raw Skill collections inevitably confounds their structural organization with varying author styles, content coverage, and implicit helper implementations. To cleanly isolate organization as an independent runtime intervention, SkillJuror transforms source artifacts into a semantically matched pair: a flat \textbf{Baseline} and a structurally reorganized \textbf{Progressive Disclosure (PD)} variant, as exemplified in Figure~\ref{fig:controlled-skill-bundle-example} to offer an intuitive understanding of their concrete archetypes.

Following established industrial Agent Skill guidance from Anthropic~\citep{anthropic2025agentskills}, the PD paradigm minimizes the agent's immediate cognitive load by serving a lightweight root document (\texttt{SKILL.md}) that routes to partitioned supporting resources only on demand. Conversely, the Baseline flattens this identical knowledge base into a single, self-contained file.

To strengthen semantic control, all generated candidate pairs undergo a rigorous two-layer verification pipeline before entering the evaluation engine:
\begin{enumerate}[leftmargin=2em, noitemsep]
    \item \textbf{Structural Adherence:} The PD variant must feature a concise root \texttt{SKILL.md}, maintain physically decoupled supporting directories, and provide explicit navigational hyperlinks within \texttt{SKILL.md}.
    \item \textbf{Knowledge Preservation:} Both variants must hold the underlying task scope, workflow obligations, hardware/software dependencies, exact numeric thresholds, and verifier contracts invariant relative to the source.
\end{enumerate}
This pipeline ensures that any downstream deviations in agent trajectories or pass rates are strictly attributable to structural layout rather than content drift.

\subsubsection{Skill-for-Skill Transformation Pipeline}

We implement construction through Skill-for-Skill transformations. Specifically, we employ an LLM agent equipped with a dedicated transformation Skill to ingest the source artifact and produce a new Skill artifact under the target organization profile. These transformations are executed as Harbor-backed construction runs, so the Baseline and PD artifacts are produced inside the same sandboxed task materialization and verifier boundary used by the later runtime study. This keeps construction scalable while avoiding manual rewriting and host-specific file-system assumptions.

For the Baseline artifact, the transform flattens the source Skill bundle into a controlled comparison anchor while preserving commands, helper contracts, script contents, schemas, warnings, thresholds, output formats, and workflow obligations. For the Progressive Disclosure variant, the transform starts from the Baseline artifact and reorganizes the same content into a short root \texttt{SKILL.md} and supporting files.

In the notation above, construction is a two-step task-local transformation:
\[
\mathcal{B}^{\mathrm{src}}_T
\xrightarrow{f_{\mathrm{base}}}
\mathcal{B}_{T,\mathrm{Base}}
\xrightarrow{f_{\mathrm{pd}}}
\mathcal{B}_{T,\mathrm{PD}}.
\]
The second step is interpreted as a constrained reorganization of the Baseline artifact, preserving the same task scope, helper contracts, constraints, and output requirements.

Constructing PD from the Baseline artifact is deliberate. If Baseline and PD were generated independently from the original source, differences in wording, coverage, or helper recovery could make it difficult to attribute observed differences specifically to organization. By treating PD as a constrained reorganization of the Baseline artifact, we reduce the opportunity for uncontrolled content changes. The PD transform may move detailed material into references or support files while preserving functional domains, concrete commands, helper implementations, task scope, and task-critical constraints.

\subsubsection{Construction Validation Mechanism}

To increase confidence that the constructed variants provide rigorous semantic control, SkillJuror implements a multi-layered verification framework rather than relying on a single, unconstrained judge model. By coupling rigid programmatic execution with flexible semantic auditing, the pipeline offsets the blind spots of code-based rules with the contextual nuance of model evaluation. This integrated pipeline sequentially enforces structural, semantic, and programmatic invariance through three progressive validation tiers:

\begin{itemize}[leftmargin=1.5em, noitemsep]
    \item \textbf{Deterministic Gating:} Fast, programmatic filters validate low-level artifact integrity, checking mechanical properties such as directory layout, file-path hygiene, resource reachability, and behavior-unit diffs.
    \item \textbf{Rubric-Based Semantic Auditing:} An automated semantic assessment scans the remaining candidates against localized rubrics to ensure the invariant preservation of task scopes, execution constraints, and input/output contracts. 
    \item \textbf{Human-in-the-Loop Override Policy:} Initial validation anomalies or flags are routed to an evidence-backed review protocol. These are systematically classified and resolved as either source-inherited idiosyncrasies, packaging artifacts, or handled via controlled source-to-target repairs.
\end{itemize}

A final pre-runtime eligibility filter drops semantic or scale outliers to insulate the downstream benchmarking from execution confounding. The resulting verified artifact pairs establish a clean, standardized foundation where downstream performance variances can be confidently attributed to structural layout rather than construction drift.

\subsection{Runtime Evidence Collection}

The runtime layer converts accepted Skill variants into matched execution traces. For each task, SkillJuror rebuilds the same non-Skill execution substrate -- instruction, workspace, environment, and verifier -- in a sandboxed execution task, then swaps only the exposed Skill artifact across conditions. This standardizes the non-Skill environment and keeps the comparison focused on whether the agent receives no task Skill, Baseline, or the Progressive Disclosure variant.

Each task--condition pair is executed with the same agent harness, model family, reasoning configuration, verifier, and timeout budget. We use multiple independent trials because agent execution is stochastic even when the task and Skill are fixed, and each trial starts from the same materialized task state and becomes one observation in $\mathcal{R}_{T,c}=\{r_{T,c,j}\}_{j=1}^{n}$. Main-run task counts, condition counts, and trial counts are specified in the experimental setup.

Every trial preserves the trajectory and verifier-side execution record needed for replay, audit, and aggregation. Given these comparable artifacts, the evaluation layer defines outcome, efficiency, paradigm-realization, and resource-routing measurements.

\subsection{Multi-Dimensional Skill Evaluation}

The evaluation layer maps each repeated trial set $\mathcal{R}_{T,c}$ into task-level summaries $A_d(T,c)$ along four dimensions: outcome utility, efficiency trade-off, paradigm realization, and resource-routing quality. The aggregation rule is tied to the evidence type: verifier outcomes summarize task completion, resource measures can be viewed per attempt or per strict pass\footnote{Throughout the main text, ``pass'' denotes a strict verifier pass, counted only when the verifier reward equals 1. Partial rewards are retained only as diagnostics in Appendix~\ref{app:metric-cost}; broad-acceptance diagnostics are reported in Appendix~\ref{app:layout-rq4}, Table~\ref{tab:app-layout-supplement}.}, and trajectory evidence summarizes process behavior.

\paragraph{Outcome Utility.} Outcome utility measures whether the agent completes the task. We keep runtime completion separate from verifier success so that execution failures are not conflated with incorrect solutions.

\paragraph{Efficiency Trade-off.} Efficiency measures the resource burden of attempting or completing a task. The per-trial view captures deployment cost across all attempts, while the per-pass view captures the yield-normalized cost of obtaining strict passes.

\paragraph{Paradigm Realization.} Paradigm realization asks whether the intended Skill organization is visible in runtime behavior. It uses trajectory attribution to test how much, when, and in what pattern the agent engages with Skill material.

\paragraph{Resource Routing Quality.} Resource Routing Quality asks whether Skill-resource access becomes useful local work. The main metric is Effective Resource Uptake (ERU), which counts a resource or helper signal only when it is consumed into observable implementation, validation, correction, or credible blocker diagnosis. ERU labels are assigned by an LLM-as-judge audit over extracted trajectory events and interpreted as bounded process evidence. Appendix~\ref{app:eru-protocol} reports the event categories, boundary rules, and aggregate count summaries used to interpret this metric.

\subsection{Task Attribute Schema}

Task-level contrasts are expected to be heterogeneous: the same resource exposure can help one task, remain locally useful but outcome-irrelevant in another, or add work in a long artifact pipeline. To support task-dependent analysis, SkillJuror attaches a compact attribute schema to each task. The schema combines source-provided contextual labels, when available, with mechanism-facing labels introduced by the analysis. Contextual labels preserve broad task provenance, while workflow type and validation type describe what the agent must operationally do and how success is accepted.

These labels give the later analyses a common vocabulary for grouping task-level contrasts. They are used to examine when additional resource uptake is paired with outcome or efficiency gains, and when it remains local process behavior that does not close the verifier contract.

\FloatBarrier

\section{Experiments}
\label{sec:experiments}

We organize the experiments around four questions that move from input control to outcomes, execution processes, and task-dependent heterogeneity:

\noindent\textbf{RQ1(Construction Reliability):} Can we reliably construct and audit semantically controlled Skill variants that instantiate different writing paradigms?

\noindent\textbf{RQ2(Outcome Effects):} How do different Skill paradigms affect task success and efficiency?

\noindent\textbf{RQ3(Process Effects):} How do different Skill paradigms affect agents' execution processes and resource-use behavior?

\noindent\textbf{RQ4(Task-Dependent Effects):} How do task properties influence the outcome and process effects of different Skill paradigms?

Together, these questions test whether multi-dimensional evaluation reveals Skill-paradigm effects that pass/fail evaluation alone would miss.

\subsection{Experimental Setup}

The primary run reported in the paper instantiates the notation above with a main task set $\mathcal{T}_{\mathrm{main}}$ of 82 SkillsBench tasks after applying pre-runtime eligibility filters for external dependencies, size-tail artifacts, and environment comparability. The main condition set is $\{\emptyset,\mathrm{Base},\mathrm{PD}\}$, corresponding to no Skill, Baseline, and Progressive Disclosure. We run $n=5$ trials for every task--condition pair, yielding $|\mathcal{T}_{\mathrm{main}}|n=410$ trials per condition for the main RQ2/RQ3 estimates. Full task accounting, exclusion rationale, and the layout-sensitivity supplement are reported in Appendices~\ref{app:task-set} and~\ref{app:layout-rq4}.

\begin{table}[!htbp]
\centering
\caption{\textbf{Runtime setup for the main comparison.}}
\label{tab:experimental-setup}
\small
\setlength{\tabcolsep}{4pt}
\begin{tabularx}{\linewidth}{p{0.25\linewidth}L}
\toprule
Setting & Main-text value \\
\midrule
Harness / backend & Codex runtime with Harbor-backed sandbox materialization. \\
Runtime model & GPT-5.4, high reasoning, fixed across conditions. \\
Primary trial grid & 82 tasks $\times$ 3 primary conditions $\times$ 5 trials = 1,230 trials. \\
Primary conditions & No Skill, Baseline, Progressive Disclosure. \\
Secondary layout checks & Origin on the 82-task set; Origin-flat on the 56-task multi-Skill subset. \\
\bottomrule
\end{tabularx}
\end{table}

Tokens/pass reports non-cached input, cache-creation input, and output tokens, excluding cached-read tokens from the display total. Cost/pass uses the same yield-normalized denominator as minutes/pass, after separating non-cached input, cache-creation input, cached-read input, and output tokens under the GPT-5.4 standard-rate assumption. Appendix~\ref{app:metric-cost} gives the pass criterion and computation protocol.


\subsection{Construction Reliability (RQ1)}

RQ1 asks whether SkillJuror can build two Skill versions that keep the same task meaning while differing in organization. This construction-reliability step establishes that RQ2--RQ4 compare organizational differences rather than drift in task scope, helper availability, command semantics or output contracts.

The construction pipeline completed both Baseline and Progressive Disclosure artifacts for all 88 construction-eligible tasks. Table~\ref{tab:rq1-construction-gates} reports the three validation tiers introduced in Section~\ref{sec:method}. Deterministic gating verified that the materialized artifacts satisfied style, path, and behavior-unit checks before runtime evaluation. Rubric-based semantic auditing then checked the accepted variants against preservation and Progressive Disclosure criteria across all 968 rubric items under a policy-aware Harbor GPT-5.4 audit, yielding 3/968 negative rubric checks. Finally, the human-in-the-loop override policy adjudicated those three cases, leaving no unresolved construction-validity issue. Appendix~\ref{app:construction-audit} reports the rubric criteria and gate outcomes.

\begin{table}[!htbp]
\centering
\caption{\textbf{Main construction-reliability summary for the controlled Skill artifacts.} The table follows the three validation tiers defined in Section~\ref{sec:method}; detailed gate outcomes are reported in Appendix~\ref{app:construction-audit}.}
\label{tab:rq1-construction-gates}
\setlength{\tabcolsep}{3pt}
\begin{tabularx}{\linewidth}{p{0.36\linewidth}L}
\toprule
Gate & Result \\
\midrule
Deterministic Gating & 88 tasks validated; semantic content preserved \\
Rubric-Based Semantic Auditing & 176 variants evaluated; 965/968 items passed automated review \\
Human-in-the-Loop Adjudication & 3 items required manual review; all accepted under documented policy\\

\bottomrule
\end{tabularx}
\end{table}

The artifacts can therefore be used as controlled inputs for runtime evaluation: the accepted variants preserve task requirements while differing in organization. RQ2--RQ4 then test whether that organizational difference changes outcomes, execution behavior, and task-dependent effects.

\rqanswer{The controlled artifacts have passed the construction-reliability protocol. Both variants were generated for all 88 eligible tasks, all deterministic gates were passed, and semantic-audit findings were resolved through evidence-based review.}

\FloatBarrier

\subsection{Outcome Effects (RQ2)}

RQ2 asks whether controlled Skill paradigms change task success and efficiency. We use verifier pass rate as the primary outcome measure. Aggregate pass rate is the fraction of trials in a condition that receive a strict verifier pass, and the headline controlled effect is the Progressive Disclosure pass rate minus the Baseline pass rate.

Efficiency has several useful denominators. The headline columns in Table~\ref{tab:rq2-outcome-efficiency} use a yield-normalized denominator: total time, display tokens, or estimated cost across all attempts divided by the number of strict passes. This metric estimates the resource burden of obtaining one strict pass under repeated attempts, so it includes failed trials in the numerator. It is thus distinct from both the mean resource use per attempted trial and the mean resource use among successful trials only; Table~\ref{tab:rq2-efficiency-views} reports all three views for time and estimated dollar cost.

\begin{table}[!htbp]
\centering
\caption{\textbf{Pass outcomes and yield-normalized per-pass efficiency for the main RQ2 comparison.} Bold condition names report the primary 82-task conditions; bold numeric entries mark the best primary-condition value in the column, with higher better for pass/rate/$\Delta$ and lower better for resource columns. Italic rows show secondary Origin and Origin-flat layout checks under the same verifier-pass criterion. $\Delta$ is relative to the comparable Baseline where applicable. Min/pass, Tokens/pass, and Cost/pass divide total resource use across all attempts by strict passes. Token/cost details and layout accounting appear in Appendices~\ref{app:metric-cost} and~\ref{app:layout-rq4}.}
\label{tab:rq2-outcome-efficiency}
\setlength{\tabcolsep}{2pt}
\begin{tabular}{llrrrrrr}
\toprule
Condition & Estimate set & Pass & Rate & $\Delta$ & Min/pass & Tokens/pass & Cost/pass \\
\midrule
\textbf{No Skill} & 82 tasks & 119/410 & 29.0\% & -53 & 29.8 & 0.34M & \$2.05 \\
\textbf{Baseline} & 82 tasks & 172/410 & 42.0\% & -- & 20.1 & 0.22M & \textbf{\$1.28} \\
\textbf{Progressive Disclosure} & 82 tasks & \textbf{189/410} & \textbf{46.1\%} & \textbf{+17} & \textbf{17.8} & \textbf{0.21M} & \$1.31 \\
\addlinespace[2pt]
\multicolumn{8}{l}{\textit{Secondary Origin and Origin-flat layout checks}} \\
\midrule
\textit{Origin} & 82 tasks & 172/410 & 42.0\% & 0 & 17.8 & 0.31M & \$1.68 \\
\textit{Origin-flat} & 56 tasks & 112/280 & 40.0\% & -- & 20.1 & 0.37M & \$1.92 \\
\bottomrule
\end{tabular}
\end{table}

\begin{table}[!htbp]
\centering
\caption{\textbf{Three efficiency views for the primary 82-task RQ2 conditions.} ``Attempt'' averages over all trials; ``success path'' averages only strict-passing trials; ``yield'' divides total resource use across all trials by strict passes, matching the Min/pass and Cost/pass columns in Table~\ref{tab:rq2-outcome-efficiency}. Bold numeric entries mark the lowest value in each column.}
\label{tab:rq2-efficiency-views}
\small
\setlength{\tabcolsep}{4pt}
\begin{tabular}{lrrrrrr}
\toprule
Condition & Attempt & Success & Yield & Attempt & Success & Yield \\
 & min/trial & path min & min/pass & cost/trial & path cost & cost/pass \\
\midrule
No Skill & 8.64 & 6.80 & 29.76 & \$0.60 & \$0.53 & \$2.05 \\
Baseline & 8.42 & 7.40 & 20.07 & \textbf{\$0.54} & \textbf{\$0.50} & \textbf{\$1.28} \\
Progressive Disclosure & \textbf{8.19} & \textbf{6.30} & \textbf{17.76} & \$0.61 & \textbf{\$0.50} & \$1.31 \\
\bottomrule
\end{tabular}
\end{table}

Table~\ref{tab:rq2-outcome-efficiency} supports three outcome-efficiency patterns. First, both Skill conditions improve strict pass rate relative to No Skill: Baseline reaches 42.0\%, and Progressive Disclosure reaches 46.1\%, compared with 29.0\% for No Skill. The controlled PD--Baseline gap is 17 additional strict passes out of 410 matched trials (+4.1\%), with a paired task-level 95\% CI half-width of $\pm$6.0\%, so the aggregate gain is  \textbf{positive but modest}.

Second, the efficiency profile shows a trade-off, not uniform dominance. Progressive Disclosure lowers yield-normalized wall-clock burden from 20.1 to 17.8 minutes per strict pass, but cost/pass remains nearly tied with Baseline (\$1.31 vs. \$1.28) partially because higher pass yield offsets higher per-attempt cost. Third, the secondary Origin and Origin-flat rows contextualize the source-layout explanation: they match the comparable Baseline pass count while requiring more displayed tokens and estimated cost per pass. Full layout accounting appears in Appendix~\ref{app:layout-rq4}.

\begin{figure}[!htbp]
\centering
\includegraphics[width=0.58\linewidth]{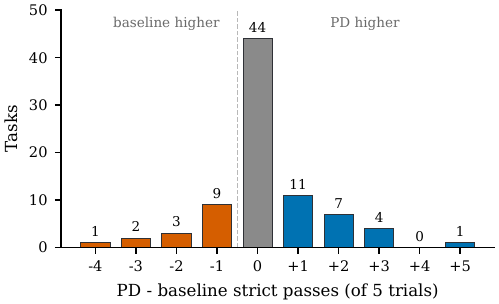}
\caption{\textbf{Per-task pass-count deltas for Progressive Disclosure relative to Baseline.} Counts cover 82 tasks with five trials per condition. Positive and negative bars show task-level heterogeneity around the aggregate gain.}
\label{fig:rq2-task-delta-distribution}
\end{figure}

The aggregate effect is positive but heterogeneous. As Figure~\ref{fig:rq2-task-delta-distribution} shows, Progressive Disclosure improves over Baseline on 23 tasks, ties on 44 tasks, and underperforms on 15 tasks under the pass metric. Most nonzero differences are one or two trials, but both positive and negative tails are present. 

This pronounced heterogeneity underscores a key insight: looking only at final pass rates gives an incomplete picture. Because the aggregate outcome gain is modest and highly task-dependent, binary success metrics fail to capture how organization actually shapes agent reasoning across different tasks. This layout change operates primarily as a behavioral intervention, altering how an agent searches for and applies knowledge throughout its execution trace, before it ever hits a verifier. This disconnect motivates the process analysis in RQ3 and the task-type analysis in RQ4, where we look beyond final endpoints to audit how structural layout reshapes runtime trajectories and resource uptake.

\rqanswer{Progressive Disclosure yields a modest aggregate pass-rate gain and faster successful completions with only a marginal cost increase, though the task-level effect remains heterogeneous.}

\FloatBarrier

\subsection{Process Effects (RQ3)}

RQ3 examines whether altering a Skill's structure modifies an agent's navigation and consumption of procedural knowledge. While pass rates capture final outcomes, the behavioral impact of the layout intervention is most transparent here, demonstrating that Progressive Disclosure fundamentally reshapes agent trajectories regardless of final task success. We operationalize this behavioral audit through two lenses: macro trajectory properties (Paradigm Realization) and granular resource routing (Resource Routing Quality).

\subsubsection{Paradigm Realization}

Table~\ref{tab:rq3-paradigm-realization} shows that the layout intervention is visible in runtime behavior. Progressive Disclosure roughly doubles Skill-step share (5.4\% to 10.8\%) and increases mean resource fanout from 1.18 to 3.85 resources per trial, while support-file and helper evidence rise sharply. These measures capture routing and access behavior. The ERU analysis below examines how often the additional resource interactions are taken up in concrete local work.

\begin{table}[!htbp]
\centering
\caption{\textbf{RQ3 Paradigm Realization signals.} Signals come from deterministic trajectory attribution for the main Baseline--Progressive Disclosure comparison. ``Fanout'' is the mean number of distinct Skill resources touched per trial. These signals describe runtime behavior changes; ERU below examines how often accessed resources are taken up in concrete work.}
\label{tab:rq3-paradigm-realization}
\small
\setlength{\tabcolsep}{2pt}
\begin{tabular}{lrrrrrr}
\toprule
Condition & Skill step & Skill token & Support read & Script & Helper & Fanout \\
 & share & share &  &  &  &  \\
\midrule
Baseline & 5.4\% & 9.0\% & 66/410 & 31/410 & 3/410 & 1.18 \\
Progressive Disclosure & 10.8\% & 11.5\% & 364/410 & 101/410 & 118/410 & 3.85 \\
\bottomrule
\end{tabular}
\end{table}

\begin{figure}[!htbp]
\centering
\includegraphics[width=0.86\linewidth]{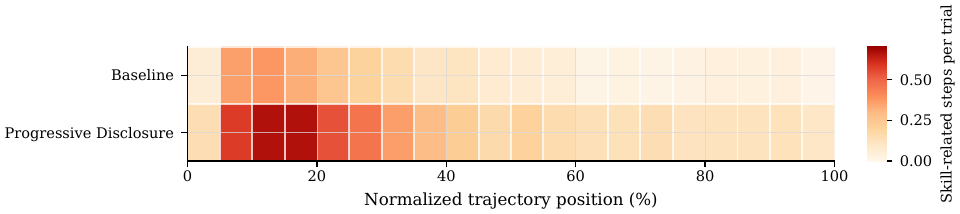}
\caption{\textbf{Skill-step timing in the Baseline--Progressive Disclosure comparison.} Columns are 5\% bins of normalized trajectory position. Color encodes absolute Skill-related steps per trial. The main visible shift is that PD carries more Skill use into middle and late trajectory phases while also increasing the absolute amount of Skill use within each phase.}
\label{fig:rq3-skill-step-heatmap}
\end{figure}

The aggregate table hides an important temporal difference. Figure~\ref{fig:rq3-skill-step-heatmap} plots where Skill-related steps occur along normalized trajectory position. Baseline is concentrated near the beginning of the trajectory, consistent with early Skill intake followed by general work. Progressive Disclosure remains early-heavy, but it spreads Skill use farther into the middle and late phases. The split across trajectory thirds makes this shift explicit:

\begin{center}
\scriptsize
\begin{tabular}{lrrr}
\toprule
Condition & Early & Middle & Late \\
\midrule
Baseline & 69.5\% & 21.9\% & 8.5\% \\
Progressive Disclosure & 59.4\% & 25.1\% & 15.5\% \\
\bottomrule
\end{tabular}
\end{center}

The distributed-skill-trial rate also rises from 20.7\% to 48.4\%. Progressive Disclosure changes the work loop. Agents do not only read a larger instruction once, but return to support resources while implementing, checking, and repairing.

\subsubsection{Resource Routing Quality}

The Paradigm Realization layer shows that Progressive Disclosure routes agents into more Skill resources, while ERU asks whether those additional accesses become usable work. An agent can scan directories, probe wrong paths, read generic background material, or invoke a helper without using the resulting signal. We therefore use ERU as a stricter Resource Routing Quality measure over extracted resource events from each trial. A resource event contributes to ERU only when the agent consumes the resource content, helper output, or diagnostic signal and turns it into observable local progress, validation, correction, or credible blocker diagnosis.

Table~\ref{tab:rq3-eru-sample} reports the ERU summary over 410 trajectories per condition. The headline pattern is exposure plus uptake intensity: Progressive Disclosure increases extracted resource events from 717 to 1,902, and mean ERU-positive uptake events per trajectory from 1.33 to 3.92. The main RQ3 signal is that agents open more files and that more of those resource signals become observable implementation, validation, correction, or diagnosis steps.

\begin{table}[!htbp]
\centering
\caption{\textbf{RQ3 Resource Routing Quality evidence.} Counts cover the full Baseline--Progressive Disclosure comparison. ``Events'' counts extracted Skill-resource events; ``Mean yes'' counts only ERU-positive events consumed into local implementation, validation, correction, or credible diagnosis.}
\label{tab:rq3-eru-sample}
\small
\setlength{\tabcolsep}{2pt}
\begin{tabular}{lrrrrrr}
\toprule
Condition & ERU rate & Events & Mean events & Mean yes & $\geq$2 yes & $\geq$5 yes \\
 & & & /traj. & /traj. & traj. & traj. \\
\midrule
Baseline & 76.0\% & 717 & 1.75 & 1.33 & 95/410 & 11/410 \\
Progressive Disclosure & 84.6\% & 1902 & 4.64 & 3.92 & 335/410 & 118/410 \\
\bottomrule
\end{tabular}
\end{table}

Task-level ERU rates remain heterogeneous, but the aggregate pattern shows that Progressive Disclosure increases resource interaction, later revisitation, and observable local uptake. The RQ3 finding reveals that the PD layout drives agents to open more support resources, revisit them later, and turn more of those signals into concrete local work. Crucially, this is where the layout intervention is most visible in the entire framework: even when final outcome translation is incomplete or blocked by task-specific constraints, the PD paradigm is behaviorally realized, thereby changing the agent's runtime reasoning and execution path. This metric is therefore best read as process evidence for how the layout changes resource use.

\rqanswer{Progressive Disclosure is reflected in runtime behavior, driving agents to access more distinct files, execute downstream revisitations, and more frequently translate those interactions into ERU-positive local uptake.}

\FloatBarrier

\subsection{Task-Dependent Effects (RQ4)}

\begin{figure}[!htbp]
\centering
\includegraphics[width=\linewidth]{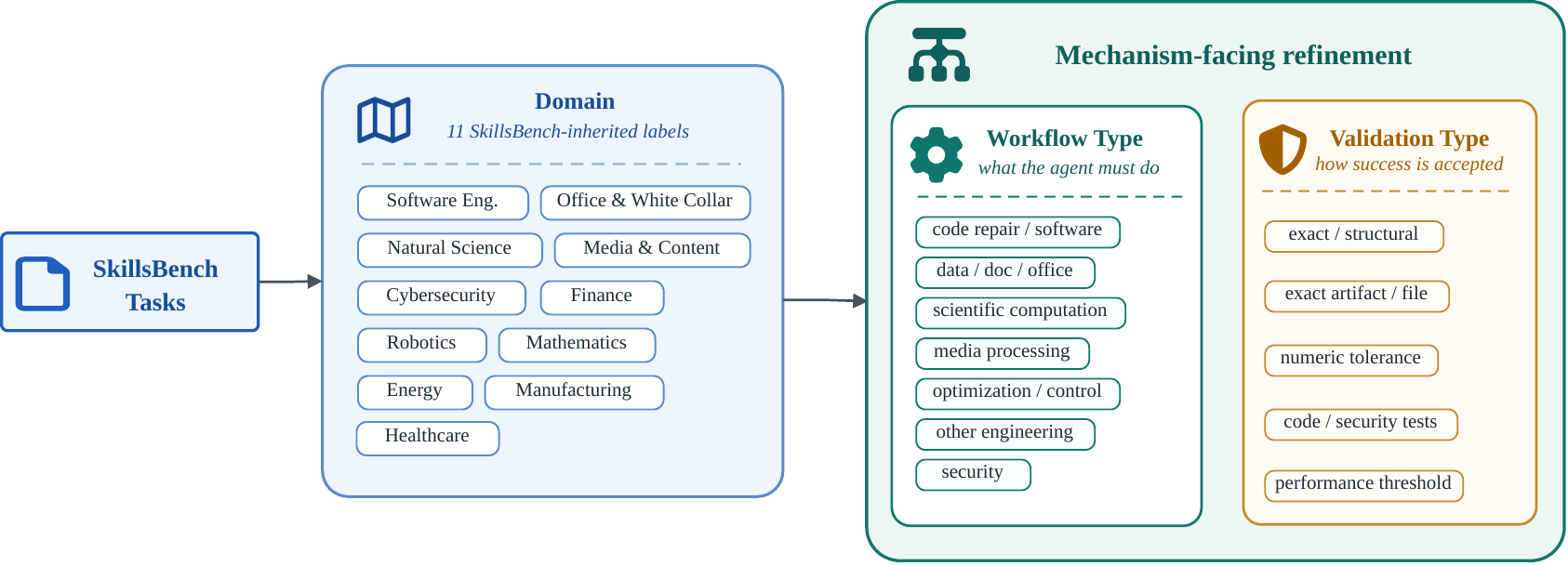}
\caption{\textbf{Task-type labeling schema for the SkillsBench instantiation.} Domain labels are inherited from the source benchmark, while workflow type and validation type are added as mechanism-facing labels for interpreting task-dependent process--outcome patterns.}
\label{fig:task-attribute-schema}
\end{figure}

RQ4 asks when the resource-use changes observed in RQ3 translate into the outcome and efficiency changes observed in RQ2. This task-level heterogeneity underscores that pass-rate metrics alone do not capture the subtle dynamics of the layout intervention, necessitating a more granular analytical lens. To isolate when and how the expanded resource routing effectively translates into macro outcome gains, we employ task attributes as analytical moderators through a dual-coordinate schema, as illustrated in Figure~\ref{fig:task-attribute-schema}.

While retaining source-provided domain labels for broad context, we introduce two mechanism-facing analysis labels to systematically audit the alignment between process indicators and verifier contracts: \emph{workflow type}, which captures the operational nature of the agent's work, and \emph{validation type}, which defines the strictness of the evaluation boundary. This integrated stratification framework permits a granular look at the experimental data based on how success is operationally achieved and accepted, thereby revealing distinct empirical translation archetypes.

\subsubsection{Type-Level Translation Patterns}

\begin{figure*}[!htbp]
\centering
\includegraphics[width=\textwidth]{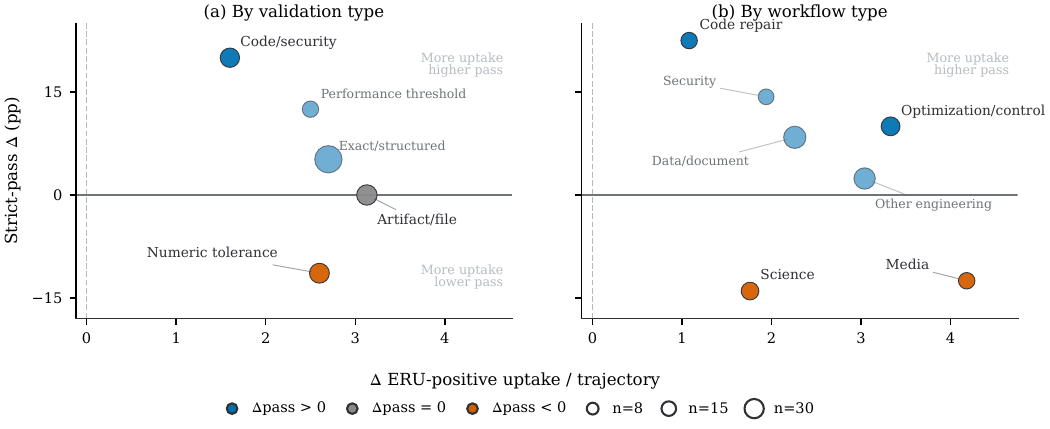}
\caption{\textbf{Process--outcome alignment across selected task strata.} Points are validation-type and workflow-type strata from the RQ4 support analysis. The x-axis shows the PD-minus-Baseline change in ERU-positive events per trajectory, and the y-axis shows the strict-pass-rate delta. Color marks the pass-delta sign, and marker area marks the number of tasks.}
\label{fig:rq4-uptake-outcome-scatter}
\end{figure*}

Figure~\ref{fig:rq4-uptake-outcome-scatter} visualizes the localized interaction between process intensity (the change in ERU-positive events) and final outcome variance across our stratified coordinates, with the highly interpretable regions clustered into four empirical translation archetypes in Table~\ref{tab:rq4-type-fingerprints}. We treat these recurring profiles as conditional diagnostics reflecting the common structural pathways through which enhanced resource routing is either processed into verifier-visible action, absorbed as redundant overhead, or decoupled from final validation requirements. Appendix~\ref{app:layout-rq4} reports the selective support strata, weak-signal rationale, and compact bridge traces behind these descriptive cases.

\begin{table}[!htbp]
\centering
\caption{\textbf{Selected RQ4 task-type empirical translation patterns.} Deltas are Progressive Disclosure minus Baseline. The table foregrounds interpretable, high-signal diagnostic trends rather than invariant causal groupings; detailed deltas and support strata are reported in Appendix~\ref{app:layout-rq4}.}
\label{tab:rq4-type-fingerprints}
\scriptsize
\setlength{\tabcolsep}{3pt}
\begin{tabularx}{\linewidth}{>{\raggedright\arraybackslash}p{0.19\linewidth}>{\raggedright\arraybackslash}p{0.20\linewidth}>{\raggedright\arraybackslash}p{0.24\linewidth}L}
\toprule
Translation pattern & Where observed & Main signal & Diagnostic mechanism reading \\
\midrule
Targeted efficiency & Code/security tests; code repair/software; security & Pass/runtime $\uparrow$; min/token per pass $\downarrow$; ERU $\uparrow$ & Routed resources become executable, verifier-visible code and test actions. \\
Uptake without success & Numeric tolerance; scientific computation; Natural Science & ERU $\uparrow$; pass/runtime $\downarrow$ & Resources support local work, but numeric and output contracts remain binding. \\
Fanout tax & Media processing; exact artifact/file validation & Fanout/helper use $\uparrow$; artifact success $\leftrightarrow$ or $\downarrow$ & Resources expand brittle exact-artifact pipelines instead of stabilizing the final output. \\
Completion with risk & Performance threshold; optimization/control & Runtime completion $\uparrow$; pass given runtime success $\downarrow$ & Resources help reach executable completion while threshold, schema, or oracle alignment remains separate. \\
\bottomrule
\end{tabularx}
\end{table}

The resulting distribution demonstrates that aggregate pass-rate variance is merely a surface-level reflection of deeper task-type constraints, which Table~\ref{tab:rq4-type-fingerprints} groups into distinct empirical profiles based on how local behavior aligns with the verifier contract. As illustrated, code and security frequently exhibit a favorable alignment where elevated process uptake directly supports executable, verifier-visible actions. Conversely, media-processing and exact-artifact domains tend to manifest a behavioral decoupling; the layout successfully prompts local resource interaction, but these expanded pathways often introduce a fanout tax without stabilizing brittle serialization requirements. Similarly, in numeric, scientific, and optimization tasks, an organizational paradigm can be locally actionable and assist the agent along a completion path, yet final success remains independently governed by precise numerical tolerances or strict oracle alignment. These stratified trends highlight that enhanced resource routing alters runtime execution pathways across the board, but its final outcome utility remains conditional on the nature of the task's evaluation boundary.

\subsubsection{Case Studies}

We next examine representative trajectories to interpret the operational mechanics behind these empirical profiles. Each case traces a specific resource signal from local behavior to intermediate action, illustrating how that execution path interacts with the verifier contract. These qualitative traces offer a descriptive diagnostic for why certain structural layouts alter outcomes, complementing the macro tendencies observed in the aggregate statistics.

\paragraph{Targeted efficiency gain.}
In \texttt{simpo-code-reproduction}, the PD layout routes the agent from the concise root skill into the task-specific \texttt{references/nlp-reproduction.md} file. The trajectory translates this reference into an executable feedback loop: the agent aligns the environment, inspects the SimPO trainer, and executes \texttt{unit\_test/unit\_test\_1.py} with \texttt{PYTHONPATH=/root/SimPO} while monitoring \texttt{/root/loss.npz}. Because the test suite provides immediate feedback, the agent efficiently operationalizes the localized instructions, converting a 0/5 Baseline pass rate into a 5/5 strict pass under PD while reducing average latency.

\paragraph{Uptake without success.}
In \texttt{exoplanet-detection-period}, the PD variant exhibits high process engagement with supporting material, but this intense local uptake fails to translate into the verifier's numeric contract. Trajectories show effective consumption of \texttt{references/preprocessing.md}, \texttt{references/box-least-squares.md}, and \texttt{references/lomb-scargle.md} to construct a multi-stage period search. However, strict passes drop from 5/5 to 3/5. Logs reveal a process-outcome decoupling: the final output file contains a logical period (\texttt{2.79991}), but it is rejected because the verifier strictly demands an oracle value of \texttt{5.35699} within a \texttt{0.01} tolerance. This diagnoses a persistent vulnerability where layouts facilitate local procedural reasoning yet remain decoupled from unexposed validation boundaries.

\paragraph{Fanout tax.}
In \texttt{video-tutorial-indexer}, modular transcription helpers appear locally useful, but the decentralized layout inadvertently expands the agent's interaction surface within a fragile media pipeline. The agent reads \texttt{references/transcription.md} and \texttt{speech-to-text/scripts/transcribe.py}, but becomes entangled in repairing helper paths and debugging slow automated speech recognition (ASR) runs. While runtime completion holds at 5/5, strict passes fall from 5/5 to 2/5 as execution steps and tokens rise substantially. This illustrates a compounding fanout overhead. When the final contract requires exact structural serialization, dispersing instructions into on-demand files can introduce path-brittleness rather than stabilizing the generated output.

\paragraph{Completion-path gain with alignment risk.}
In \texttt{manufacturing-equipment-maintenance}, the PD layout lowers the entry-level cognitive load to establish a runnable compliance workflow where Baseline uniformly fails. The concise root directs the agent to \texttt{references/compliance-computation.md}, which successfully guides the implementation of ramp computations, TAL interpolations, and thermocouple tie-breaking. Although runtime completion rises from 0/5 to 5/5 and strict passes recover to 3/5, failure logs expose the remaining boundary clearly: one run outputs a rounded value of \texttt{1.91} where the oracle mandates \texttt{1.9}, and another violates the specific ordering for \texttt{runner\_up\_run\_ids}. This profile demonstrates that a paradigm can optimize workflow execution paths, yet final success remains independently gated by rigid output schemas.

These case studies provide descriptive reference for interpreting how task attributes shape the translation of local behavior into verifier-visible outcomes. The granular trajectory evidence suggests that the final efficacy of a Skill writing paradigm remains substantially shaped by the structural properties and feedback density of the task's evaluation boundary, highlighting the qualitative variance that macro pass rates hide.

\rqanswer{Task properties govern the alignment between process and outcome patterns. In contexts with actionable, well-defined targets, increased local uptake frequently manifests as pass-rate gains, whereas tasks bound by rigid numeric conventions or exact output schemas are closely correlated with a process–outcome disconnect.}

\FloatBarrier

\section{Discussion}
\label{sec:discussion}

The evaluation of SkillJuror demonstrates that Skill organization operates as an active runtime intervention rather than passive documentation. By holding task knowledge constant, the shift to Progressive Disclosure (PD) may alter the agent's trajectory from a single-shot ``read-and-execute'' workflow to an iterative ``implement-verify-repair'' loop. The concentrated late-stage resource interactions indicate that agents treat modular, offloaded references as on-demand assets for debugging and error recovery rather than upfront context. 

This behavioral shift provides direct engineering implications for both Skill authors and harness designers. First, modularity must prioritize actionable routing over aesthetic brevity, treating Skill loading and layout as dynamic runtime control surfaces rather than static formatting. The root \texttt{SKILL.md} should decouple high-level strategy from low-level details, functioning strictly as a lean entry point while offloading granular constraints, templates, and specifications into supporting files.  Specifying exactly \emph{when} and \emph{why} to access a file anchors the agent’s tool-use loop, prompting resource invocation precisely when execution friction occurs.

Practically, the observed heterogeneity suggests that PD is not a universal solution, requiring deployment systems to match layout paradigms to the task's validation strictness. Designers should consider PD in exploration-heavy domains (e.g., code repair, system diagnostics) where tasks involve fluid environments and success depends on surviving runtime errors through late-stage troubleshooting. Conversely, flat layouts may be preferable for constraint-bound domains (e.g., scientific computation, strict schema serialization) where success hinges on rigid numeric thresholds or exact formatting. In these fixed-boundary tasks, distributing constraints across files adds a fanout tax and risks omission as keeping information centralized may reduce the risk that critical limits are missed.

\section{Limitations}

This study isolates one writing-paradigm contrast: Progressive Disclosure versus Baseline. SkillJuror is designed to evaluate broader Skill-writing interventions. The empirical scope of this paper is the controlled comparison reported here, rather than scriptization, metadata design, multi-Skill modularization, or other authoring paradigms. The controlled variants also provide auditable semantic control as opposed to a formal proof of equivalence. Deterministic gates, behavior-unit diffs, rubric review, and an independent GPT-5.4 construction audit reduce construction drift, while subtle differences in wording, emphasis, or recoverability of task constraints may remain.

The process analyses have their own boundary. ERU and bridge traces rely on LLM-assisted semantic judgment and are best interpreted as bounded mechanism evidence rather than human-ground-truth annotation or causal mediation estimates. Generalization is also limited by the evaluation substrate: the main run uses SkillsBench tasks, Harbor-backed sandbox execution, one primary agent/model configuration, and five trials per task--condition pair. Marketplace Skills, organization-internal workflows, live APIs, human-evaluated tasks, and different agent platforms may exhibit different reuse patterns, failure modes, and sensitivity to Skill organization.

\section{Conclusion}

SkillJuror reframes Agent Skill evaluation by shifting the focus from skill availability alone to controlled structural comparison. Our study shows that how procedural knowledge is organized substantially reshapes an agent's runtime trajectory, even when task semantics are held fixed under the construction audit. Effective skill writing is therefore not merely about brevity, but about \emph{actionable routing}—transforming passive instructions into conditional guidance that actively supports execution, validation, and error repair. Ultimately, evaluating agents primarily through final pass/fail outcomes can obscure the mechanics of their success. Future benchmarks should pair verifier outcomes with granular trajectory evidence, because structural layout reshapes how an agent searches and applies knowledge before those effects appear in aggregate outcome metrics.

\bibliography{main}
\bibliographystyle{rlc}

\clearpage
\appendix
\begin{center}
{\Large\bf Appendix}
\end{center}
\vspace{1em}

\section{Task Set and Exclusions}
\label{app:task-set}

Table~\ref{tab:app-task-accounting} records the task-count relationships used in the experiments. The construction protocol succeeded on 88 construction-eligible tasks. The main runtime comparison uses 82 of these tasks and excludes six tasks before runtime comparison because their artifacts or environments would make the controlled estimate unreliable.

\begin{table}[!htbp]
\centering
\caption{\textbf{Task-set accounting for the controlled runtime study.}}
\label{tab:app-task-accounting}
\small
\setlength{\tabcolsep}{4pt}
\begin{tabularx}{\linewidth}{p{0.24\linewidth}rL}
\toprule
Set & Count & Role \\
\midrule
Construction-eligible tasks & 88 & Source tasks for Baseline and Progressive Disclosure construction. \\
Main runtime tasks & 82 & Main RQ2/RQ3 task set after pre-run exclusions. \\
Main trials & 1,230 & 82 tasks $\times$ 3 conditions $\times$ 5 trials. \\
Layout-sensitivity tasks & 56 & Aligned multi-skill supplement for Origin/Origin-flat; not merged into the 82-task main estimate. \\
\bottomrule
\end{tabularx}
\end{table}

\begin{table}[!htbp]
\centering
\caption{\textbf{Main-run exclusions from the 88 construction-eligible tasks.}}
\label{tab:app-exclusions}
\footnotesize
\setlength{\tabcolsep}{4pt}
\begin{tabularx}{\linewidth}{p{0.28\linewidth}p{0.23\linewidth}L}
\toprule
Excluded task & Category & Rationale \\
\midrule
\texttt{pg-essay-to-audiobook} & External credential & Verifier requires an external speech-to-text API credential. \\
\texttt{mhc-layer-impl} & External credential & Task semantics require Modal A100 training credentials; removing verifier preflight would not make the task runnable. \\
\texttt{organize-messy-files} & Extreme Baseline artifact & Baseline exceeds the main-run token-tail cutoff (100k tokens) and would dominate context/cost estimates. \\
\texttt{exceltable-in-ppt} & Extreme Baseline artifact & Same Baseline-token outlier rule. \\
\texttt{pptx-reference-formatting} & Extreme Baseline artifact & Same Baseline-token outlier rule. \\
\texttt{lean4-proof} & Extreme Baseline artifact & Same Baseline-token outlier rule. \\
\bottomrule
\end{tabularx}
\end{table}

The result records the 82 main-run task IDs and the aligned 56-task layout supplement. The layout supplement excludes \texttt{trend-anomaly-causal-inference} because that task is outside the aligned multi-Skill subset.

\section{Construction Validity Audit}
\label{app:construction-audit}

The construction audit evidence comes from the policy-aware Harbor GPT-5.4 audit. Table~\ref{tab:app-rubric-items} gives the rubric criteria used for controlled-variant acceptance. Table~\ref{tab:app-detailed-gates} records the deterministic construction gates and semantic-audit outcomes behind the compact main-text summary.

\begin{table}[!htbp]
\centering
\caption{\textbf{Rubric items used for controlled-variant acceptance.} Each item is evaluated with a \textsc{yes}/\textsc{no} label and one concrete evidence sentence. For semantic drift, \textsc{yes} means no problematic drift was found for that item.}
\label{tab:app-rubric-items}
\scriptsize
\setlength{\tabcolsep}{4pt}
\begin{tabularx}{\linewidth}{p{0.10\linewidth}p{0.18\linewidth}L}
\toprule
Item & Layer & Acceptance criterion \\
\midrule
PD-1 & Progressive Disclosure & \texttt{SKILL.md} is at most 300 lines, so the root can function as a concise entry point. \\
PD-2 & Progressive Disclosure & The Skill contains at least one independent non-code supporting content file or directory outside \texttt{SKILL.md}. Documentation, references, examples, templates, assets, and other on-demand non-script content count; transform metadata, generated metadata files, rubric JSON, executable scripts, and helper code do not count by themselves. \\
PD-3 & Progressive Disclosure & \texttt{SKILL.md} explicitly names each independent supporting content file or directory and explains when or why to load it. This item can be accepted only when PD-2 is accepted. \\
\addlinespace[2pt]
SD-1 & Semantic drift & The candidate preserves the source Skill's main task scope and all major functional domains. \\
SD-2 & Semantic drift & The candidate preserves required inputs, outputs, workflow obligations, and success criteria. \\
SD-3 & Semantic drift & The candidate preserves important warnings, caveats, constraints, and edge-case handling. \\
SD-4 & Semantic drift & The candidate avoids contradictory instructions, unsupported new requirements, and materially different assumptions. \\
\bottomrule
\end{tabularx}
\end{table}

The rubric boundary is semantic rather than lexical. File movement, a shorter \texttt{SKILL.md}, and Progressive Disclosure references are not penalized when the moved content remains present and navigable. Conversely, the review penalizes missing source capabilities, vague replacements for concrete constraints, incompatible API or command changes, and new mandatory dependencies, paths, tools, data assumptions, or runtime steps unless they are purely organizational and do not change task behavior. For multi-Skill sources, the comparison target is the union of the source Skills rather than a single root file.

\begin{table}[!htbp]
\centering
\caption{\textbf{Detailed construction-reliability gates behind the compact RQ1 summary.}}
\label{tab:app-detailed-gates}
\scriptsize
\setlength{\tabcolsep}{4pt}
\begin{tabularx}{\linewidth}{p{0.30\linewidth}p{0.20\linewidth}p{0.18\linewidth}L}
\toprule
Gate & Scope & Result & Audit interpretation \\
\midrule
Baseline conversion & 88 tasks & 88/88 success & Baseline materialized for all eligible tasks. \\
Progressive Disclosure conversion & 88 tasks & 88/88 success & PD variant materialized for all eligible tasks. \\
Baseline style audit & 88 tasks & No affected tasks & No archive, provenance, or style pollution found. \\
Baseline behavior diff & 88 reports & No high-risk content loss & No high-risk behavior-unit loss from Origin to Baseline. \\
PD path hygiene & 88 tasks & All tasks compliant & Path and support-file hygiene passed after review. \\
PD behavior diff & 88 reports & No high-risk content loss & No high-risk behavior-unit loss from Baseline to PD. \\
Baseline semantic-drift rubric & 88 skills & All items validated & Evidence review accepted Baseline semantics under the drift rubric. \\
PD structural + semantic rubric & 88 skills & All items validated & Evidence review accepted PD structure and semantic preservation under the rubric. \\
Harbor GPT-5.4 audit & 176 variants / 968 items & All items validated after review & Full audit covered all rubric items. Baseline boundary cases were reviewed under the documented packaging policy. \\
\bottomrule
\end{tabularx}
\end{table}

\begin{table}[!htbp]
\centering
\caption{\textbf{Policy-aware Harbor GPT-5.4 construction audit.} Negative checks are cases that required human-in-the-loop adjudication before runtime evaluation.}
\label{tab:app-gpt54-audit}
\scriptsize
\setlength{\tabcolsep}{3pt}
\begin{tabularx}{\linewidth}{p{0.30\linewidth}p{0.24\linewidth}p{0.18\linewidth}L}
\toprule
Audit stage & Scope & Result & Disposition \\
\midrule
Policy-aware GPT-5.4 rubric audit & 176 variants / 968 items & 3/968 checks required review & The final audit covered all Baseline and Progressive Disclosure variants under the documented semantic-preservation and packaging policy. \\
Human-in-the-loop adjudication & 3 negative checks required adjudication & All items validated & The remaining cases were accepted after documented review when source scope, helper contracts, trigger intent, and workflow obligations were preserved. \\
\bottomrule
\end{tabularx}
\end{table}

\section{Metric and Cost Computation}
\label{app:metric-cost}

Strict pass is defined as \texttt{reward == 1}. Broad acceptance and partial rewards are retained as diagnostics but are not counted in the main strict-pass numerator.

Per-pass metrics divide total resource use across all attempts by the number of strict passes in the same condition. They are yield-normalized resource burdens, not averages over successful trials only:
\[
\text{minutes/pass} = \frac{\sum_i \text{duration}_i/60000}{\sum_i \mathbb{1}[\text{reward}_i=1]}.
\]
Display tokens/pass excludes cached-read tokens:
\[
\text{display tokens} = \text{noncached input} + \text{cache creation input} + \text{output}.
\]
The runtime logs record cached-read tokens inside \texttt{input\_tokens}, so we separate them before display and billing:
\[
\begin{aligned}
\text{noncached input} =
\max(&0,\texttt{input\_tokens}\\
&-\texttt{cache\_creation\_input\_tokens}
-\texttt{cache\_read\_input\_tokens}).
\end{aligned}
\]
Estimated cost is computed using GPT-5.4 standard-rate assumptions.\footnote{OpenAI API pricing, \url{https://openai.com/api/pricing/}, accessed June 4, 2026.}
\[
\begin{aligned}
\text{cost} ={}&
2.50\frac{\text{noncached input}+\text{cache creation input}}{10^6}\\
&+0.25\frac{\text{cache read input}}{10^6}
+15.00\frac{\text{output}}{10^6}.
\end{aligned}
\]
The reported cost/pass column is then:
\[
\text{cost/pass} = \frac{\sum_i \text{cost}_i}{\sum_i \mathbb{1}[\text{reward}_i=1]}.
\]
This cache-read separation yields the reported cost/pass values under the stated pricing assumptions.

Condition-level strict-pass intervals and paired PD--Baseline deltas are computed over the main 82-task set. The main text reports condition-level 95\% half-widths and the paired task-clustered 95\% half-width for the PD--Baseline strict-pass delta.

\section{ERU Protocol and Bridge Audits}
\label{app:eru-protocol}

The main text defines ERU as the Resource Routing Quality metric. This appendix reports the labeling boundary rules and aggregate count tables used to interpret that metric. Positive labels require observable use of a resource signal in the current task phase, such as applying a referenced contract, using helper output to patch or validate work, or treating an error signal as a credible blocker diagnosis. Negative labels cover access without consumption, including directory browsing, failed path probes, alias checks, and generic background reads that do not change the local work. Unknown labels are reserved for events whose trajectory context is insufficient to decide consumption.

\begin{table}[!htbp]
\centering
\caption{\textbf{ERU raw counts.} Mean columns divide by all 410 trajectories in each condition. The ERU-rate denominator excludes unknown events.}
\label{tab:app-eru-raw}
\small
\setlength{\tabcolsep}{4pt}
\begin{tabular}{lrrrrrrrr}
\toprule
Condition & Traj. & Events & Yes & No & Unk. & ERU rate & Mean events & Mean yes \\
\midrule
Baseline & 410 & 717 & 545 & 172 & 0 & 76.0\% & 1.75 & 1.33 \\
Progressive Disclosure & 410 & 1902 & 1609 & 292 & 1 & 84.6\% & 4.64 & 3.92 \\
\bottomrule
\end{tabular}
\end{table}

\begin{table}[!htbp]
\centering
\caption{\textbf{ERU-positive trajectory incidence.} Counts are trajectory counts out of 410 in the aggregation.}
\label{tab:app-eru-incidence}
\small
\setlength{\tabcolsep}{3pt}
\begin{tabular}{lrrr}
\toprule
Condition & $\geq$1 yes traj. & $\geq$2 yes traj. & $\geq$5 yes traj. \\
\midrule
Baseline & 332/410 & 95/410 & 11/410 \\
Progressive Disclosure & 386/410 & 335/410 & 118/410 \\
\bottomrule
\end{tabular}
\end{table}

\begin{table}[!htbp]
\centering
\caption{\textbf{ERU reliability and boundary checks.} The table summarizes coverage and boundary checks for the ERU labeling protocol.}
\label{tab:app-eru-reliability}
\scriptsize
\setlength{\tabcolsep}{3pt}
\begin{tabularx}{\linewidth}{p{0.22\linewidth}p{0.22\linewidth}p{0.18\linewidth}L}
\toprule
Audit check & Scope & Result & Claim boundary \\
\midrule
Labeling coverage & ERU pass & 807/807 non-empty calls & Supports full-condition ERU coverage for the labeled event set. \\
Boundary-case check & Known routing-boundary cases & Passed recorded checks & Checks that ERU-positive labels require observable consumption rather than access-only events. \\
Bridge-example check & Representative bridge boundary cases & Recorded check & Documents the prompt-slicing boundary used for bridge examples. \\
Human/advisor audit status & Human reliability annotation & Out of scope & ERU is mechanism evidence; this version reports audit-boundary checks rather than a human-ground-truth reliability rate. \\
\bottomrule
\end{tabularx}
\end{table}

These checks clarify the scope of the ERU and bridge analyses. The current manuscript treats ERU as mechanism evidence and leaves human-grounded reliability estimation to future audit work.

For RQ4 case interpretation, the bridge schema records a compact chain:
\[
\text{resource signal} \rightarrow \text{consumed action} \rightarrow \text{local effect} \rightarrow \text{outcome/efficiency link}.
\]

\section{Layout Sensitivity and RQ4 Support}
\label{app:layout-rq4}

\subsection{Source-Layout Supplement}
\label{app:source-layout-supplement}

The layout supplement records the secondary diagnostic rows summarized in the RQ2 table. It is not part of the 82-task primary estimate. The first audit view uses broad acceptance on the aligned 56-task multi-Skill subset. In this supplement, Origin and Origin-flat each reach 126/280 broad passes, only +2 over the comparable Baseline, while Progressive Disclosure reaches 141/280. The strict column in Table~\ref{tab:app-layout-supplement} shows why the main text uses the stricter RQ2 denominator: the source-derived layouts tie the comparable Baseline under strict reward.

\begin{table}[!htbp]
\centering
\caption{\textbf{Layout-sensitivity supplement for the aligned 56-task multi-Skill subset.} Broad acceptance is retained as a diagnostic view; strict reward remains the RQ2 primary outcome.}
\label{tab:app-layout-supplement}
\scriptsize
\setlength{\tabcolsep}{3pt}
\begin{tabularx}{\linewidth}{p{0.19\linewidth}rrrrL}
\toprule
Condition & Broad pass & Broad rate & Strict pass & Runtime success & Interpretation \\
\midrule
Baseline & 124/280 & 44.3\% & 112/280 & 219/280 & Comparable controlled anchor on the same 56 tasks. \\
Origin & 126/280 & 45.0\% & 112/280 & 246/280 & Original SkillsBench source layout. \\
Origin-flat & 126/280 & 45.0\% & 112/280 & 235/280 & Flattened original source layout. \\
Progressive Disclosure & 141/280 & 50.4\% & 126/280 & 228/280 & Controlled PD layout on the same subset. \\
\bottomrule
\end{tabularx}
\end{table}
\FloatBarrier

\subsection{Origin-Flat Layout Ablation}
\label{app:origin-flat-ablation}

Origin-flat is a flattening ablation of the original SkillsBench layout. It is directly observed on the 56 tasks whose source layout contains multiple Skills. For the remaining 26 single-Skill tasks, flattening removes no source-layout distinction: the single source Skill collapses to the same accounting role as Baseline. Table~\ref{tab:app-origin-flat-accounting} therefore reports two views. The observed 56-task view is the direct ablation. The 82-task imputed view instead combines the observed Origin-flat rows with Baseline rows on the 26 single-Skill tasks. This imputed view is an audit of layout accounting, not a new primary condition.

\begin{table}[!htbp]
\centering
\caption{\textbf{Origin-flat accounting views under strict reward.} Tokens/pass and Cost/pass use the same cache-adjusted protocol as Table~\ref{tab:rq2-outcome-efficiency}.}
\label{tab:app-origin-flat-accounting}
\scriptsize
\setlength{\tabcolsep}{3pt}
\begin{tabularx}{\linewidth}{llrrrrrL}
\toprule
Condition & Accounting set & Strict pass & Runtime & Min/pass & Tokens/pass & Cost/pass & Role \\
\midrule
Baseline & 56 observed & 112/280 & 219/280 & 21.7 & 0.25M & \$1.52 & Comparable anchor for the multi-Skill subset. \\
Origin-flat & 56 observed & 112/280 & 235/280 & 20.1 & 0.37M & \$1.92 & Direct flattening ablation used in Table~\ref{tab:rq2-outcome-efficiency}. \\
Baseline & 82 primary & 172/410 & 313/410 & 20.1 & 0.22M & \$1.28 & Main Baseline. \\
Origin-flat & 82 imputed & 172/410 & 329/410 & 19.1 & 0.30M & \$1.54 & Observed 56-task Origin-flat rows plus Baseline values for the 26 single-Skill tasks. \\
\bottomrule
\end{tabularx}
\end{table}
\FloatBarrier

The two views lead to the same substantive conclusion. On the observed 56-task subset, Origin-flat ties the comparable Baseline in strict passes but uses more displayed tokens and estimated cost per strict pass. In the 82-task accounting, Origin-flat again ties Baseline in strict pass count (172/410). Flattening the source layout alone does not explain the strict-pass gain attributed to Progressive Disclosure in the main RQ2 comparison.

The selected RQ4 translation patterns are supported by high-signal strata where outcome, efficiency, and process metrics move together. Low-$n$ domains and mixed data/document/finance groups are not used as main claims because their effects are useful diagnostics but less stable as population summaries.

\begin{table}[!htbp]
\centering
\caption{\textbf{Selected RQ4 support strata used to motivate the four main translation patterns.}}
\label{tab:app-rq4-support}
\scriptsize
\setlength{\tabcolsep}{3pt}
\begin{tabularx}{\linewidth}{p{0.22\linewidth}rL}
\toprule
Support stratum & $n$ & Why selected \\
\midrule
Code/security tests & 13 & Pass +20.0\%, runtime +15.4\%, minutes/pass -11.8, with process and outcome metrics moving in the same direction. \\
Numeric tolerance & 14 & Pass -11.4\% despite higher uptake evidence, separating local uptake from final numeric alignment. \\
Media processing & 8 & Pass -12.5\% and minutes/pass +129 with large fanout/helper increases. \\
Optimization/control & 12 & Runtime +25.0\% and higher uptake evidence with remaining pass/runtime alignment risk. \\
\bottomrule
\end{tabularx}
\end{table}

\begin{table}[!htbp]
\centering
\caption{\textbf{Exploratory strata not foregrounded in the RQ4 main text.}}
\label{tab:app-rq4-not-selected}
\scriptsize
\setlength{\tabcolsep}{3pt}
\begin{tabularx}{\linewidth}{p{0.26\linewidth}L}
\toprule
Not foregrounded & Reason \\
\midrule
Data/document/office and finance & Often show conditional gains and high uptake, but runtime and cost are mixed; better treated as process/outcome trade-off evidence. \\
Small-$n$ domains & Useful for audit completeness but too small for headline moderator claims. \\
All two-dimensional combinations & The cross-product of domain, workflow, validation, artifact, and cost variables is exploratory and would overstate precision. \\
\bottomrule
\end{tabularx}
\end{table}

\end{document}